\documentclass[10pt,twocolumn,letterpaper]{article}

\usepackage{cvpr}
\usepackage{times}
\usepackage{epsfig}
\usepackage{graphicx}
\usepackage{amsmath}
\usepackage{float}
\usepackage{amssymb}
\usepackage{subfig}
\usepackage[linesnumbered,ruled,vlined]{algorithm2e}

\usepackage[breaklinks=true,bookmarks=false]{hyperref}

\cvprfinalcopy
\def\cvprPaperID{****}

\newcommand{\kicking}{\textit{`kicking'}}
\newcommand{\swinging}{\textit{`swinging-bench'}}
\newcommand{\MIL}{\textsc{MIL}}
\newcommand{\RGB}{\textsc{RGB}}
\newcommand{\WSL}{\textsc{WSL}}
\newcommand{\CNN}{\textsc{CNN}}
\newcommand{\ADT}{\textsc{ADT}}
\newcommand{\ATM}{\textsc{ATM}}
\newcommand{\HVC}{\textsc{HVC}}
\newcommand{\IOU}{\textsc{IOU}}
\newcommand{\mAP}{\textsc{mAP}}

\newcommand{\AP}{\textsc{AP}}
\newcommand{\mSERO}{\textsc{mSERO}}

\begin{document}

\title{Spatio-Temporal Action Localization in a Weakly Supervised Setting}

\author{Kurt Degiorgio\\
Oxford Brookes University\\
{\tt\small degiorgiokurt@outlook.com}
\and
Fabio Cuzzolin\\
Oxford Brookes University\\
{\tt\small  fabio.cuzzolin@brookes.ac.uk}}

\maketitle

\begin{abstract}

Enabling computational systems with the ability to localize actions in video-based content has manifold applications. Traditionally, such a problem is approached in a fully-supervised setting where video-clips with complete frame-by-frame annotations around the actions of interest are provided for training. However, the data requirements needed to achieve adequate generalization in this setting is prohibitive. In this work, we circumvent this issue by casting the problem in a weakly supervised setting, i.e., by considering videos as labelled `sets' of unlabelled video segments. Firstly, we apply unsupervised segmentation to take advantage of the elementary structure of each video. Subsequently, a convolutional neural network is used to extract \RGB{} features from the resulting video segments. Finally, Multiple Instance Learning (\MIL{}) is employed to predict labels at the video segment level, thus inherently performing spatio-temporal action detection. In contrast to previous work, we make use of a different \MIL{} formulation in which the label of each video segment is continuous rather then discrete, making the resulting optimization function tractable. Additionally, we utilize a set splitting technique for regularization. Experimental results considering multiple performance indicators on the UCF-Sports data-set support the effectiveness of our approach.  

\end{abstract}

\section{Introduction}

Localising and classifying human actions in video-clips is a hard problem. This can be attributed to the sheer variety of different scenarios under which such actions may be performed. Traditionally a fully-supervised paradigm is used to tackle both the localization and classification of human actions. This is problematic, especially in a deep learning setting, as it is intractable to manually label spatio-temporal actions in millions of videos. Tools that automate the collection and annotation of  data are frequently used to alleviate this problem  \cite{springerlink:10.1007/s11263-012-0564-1,bianco2015interactive}. While  effective, such tools still rely on human intervention, limiting the range of action classes that can be learned. Recently, researchers have formulated the aforementioned problem under a \emph{Weakly Supervised Learning} (\WSL{})  \cite{sapienzalearning,nguyen2009weakly,pandey2011scene,hartmann2012weakly} setting, where the extensive data annotation requirements demanded in a fully supervised setting is significantly alleviated. Specifically, \WSL{} only requires one label designating the presence of a given action per video, rather than one label for every video frame or segment.  
Under weakly labelled conditions, classification functions must be learned by correlating negative and positive video instances of a given action. The process is naturally more challenging than in a fully supervised setting, as the decision of which function to learn is more ambiguous. Should a learner recognize instances were a human is \textit{`kicking a ball'}? or should it detect \kicking{} in general,  regardless of what is being kicked or who is doing the \kicking? In \WSL, the learner will  make this choice according to what is present in the extracted features. More specifically, if a majority of the positive videos depict a human body kicking a ball, then this is what the model will learn. This can be problematic as there are no labels to guide the algorithm towards the desired function, the one mapping the \kicking{} action to the \kicking{} label.
It may instead learn a function that maps the label -- \kicking-- to an artifact that is, by mere accident, prevalent in all the positive videos in the training set. Such issues notwithstanding, significant progress has been made.

The training algorithm used by the vast majority of previous related works is a formulation of \emph{Multiple Instance Learning} (\MIL). While effective, \MIL{} leads to a mixed integer programming problem that has to be solved heuristically. Moreover, formulations under this setting make it difficult to integrate prior information. 

\subsection{Contribution}

For these reasons in this paper we adopt a different approach that allows the latent label of each proposal to be continuous instead of discrete. Additionally, we make use of a video-splitting technique aimed to reduce ambiguity between proposals, thus ameliorating localization performance.
With regards to proposal generation (a crucial step in \MIL) we evaluate three different techniques that aim to generate video segments in the form of temporally-consistent \emph{action-tubes} \cite{Gkioxari_2015_CVPR}. 

Subsequently, we take advantage of a standard \textit{Convolutional Neural Network} (\CNN) architecture to extract deep features from each action tube. 

To the best of our knowledge, this holistic approach of using relaxed \MIL{} constraints in conjunction with video splitting is a novel method for localizing spatio-temporal actions under weakly labelled conditions. 

\subsection{Paper outline}

The remaining sections of this paper are structured as follows. Section II gives a concise overview of the relevant literature. Sections III and IV formalise the proposed methodology and the experimental setup, respectively. The paper concludes by presenting and discussing the observed empirical results on the UCF-Sports data-set. 

\section{Related Work}

Three distinct problems need to be tackled in order to successfully localize spatio-temporal actions in test videos: (i) feature extraction, (ii) video segment representation, and (iii) the design of algorithms that can learn to generalize over the observed features. 

The majority of state-of-the-art methods in video based action localisation employ \CNN{}s along with a temporal association algorithm for localizing actions in a fully-supervised setting \cite{Gkioxari_2015_CVPR,weinzaepfel2015learning,kang2016object,singh2017online,saha2017amtnet,behl2017incremental, DBLP:journals/corr/SahaSSTC16,DBLP:journals/corr/abs-1808-07712}. Such approaches demand per-frame bounding box annotation, which is expensive, and are hence restricted to relatively small data-sets with a limited number of action classes. This has instigated a need for a framework that can leverage the descriptive power of \CNN{}s without the expensive space-time annotations that is demanded by fully-supervised learning. Relevant research is being conducted for object detection in images \cite{DBLP:journals/corr/SongGJMHD14,NIPS2014_5284, 7298968,cinbis2015weakly,bilen2014weakly,ren2016weakly}, but the topic is still surprisingly rather unexplored in the video domain. 

Traditionally, \textit{Bag-of-features} (\textsc{BOF}) encoding has been the representation technique of choice. This method clusters space-time features to build a visual vocabulary from the entire training set of videos. Features are commonly extracted based on shape (\textsc{HOG}, etc.) or motion (e.g., optical flow) \cite{poppe2010survey}. Sapienza \textit{et al}.  \cite{sapienzalearning} have shown that this approach is inherently flawed, since the resulting feature set is not sufficiently descriptive of the action class in question. In an attempt to address this issue, \cite{sapienzalearning} divides each video into a number of sub-volumes determined by a rigged spatio-temporal grid. \textsc{BOF} or Fisher vectors are employed to describe each sub-volume by extracting and grouping features through a technique developed by \cite{wang2011action}, termed \textit{Dense Trajectory Features}. The latter is a significant improvement over previous, and now largely outdated, approaches based on interest point detectors (e.g. \textsc{3D-SIFT}, \textsc{HOG3D}, \textsc{SURF}, and so on).
After associating a video with a `bag' of video sub-volumes (a step termed \emph{proposal generation}), \cite{sapienzalearning} casts the localization problem in a weakly supervised learning framework, to select sub-volumes that best characterize the action class under consideration. Subdividing the video clip into sub-volumes and learning solely from them allows them to mitigate the issue with extraneous features. While effective, this approach is still not optimal as the grid is rigid, and cannot be made too fine because of complexity issues. For this reason, pixel-wise video segmentation, such as the one utilized by \cite{tang2013discriminative}, has the potential to be more effective. Furthermore, deep learning has shown that learning representations automatically through the use of Deep Neural Networks can outclass all (`handcrafted') representation techniques based on manual feature engineering \cite{Girshick_2014_CVPR,Gkioxari_2015_CVPR}.

Following this intuition,
Tang \textit{et al}. \cite{tang2013discriminative} annotate actions using a pixel mask by exploiting a video segmentation method conceptualised by  \cite{grundmann2010efficient} and \cite{felzenszwalb2010object}. Inter-class information is leveraged to train a WSL algorithm that is capable of localizing actions. \cite{sultani2016automatic} adopts a markedly distinctive approach. Specifically, a large number of spatio-temporal action proposals are generated using dense trajectory features. This set is significantly trimmed by exploiting motion and saliency-based information. Subsequently, a graph is constructed between proposals across different videos and actions are localized by finding maximum cliques in the resulting graph. 

The approach proposed in this paper is most similar to that of \cite{tang2013discriminative}. The novelty element in our work is the fact that we use a \emph{probabilistic \MIL{} formulation} to generalize over new features, in conjunction with a video splitting technique applied at training time, that allows us to reduce ambiguity and ameliorate generalization performance.

\begin{figure*}
\centering
\includegraphics[width=7in]{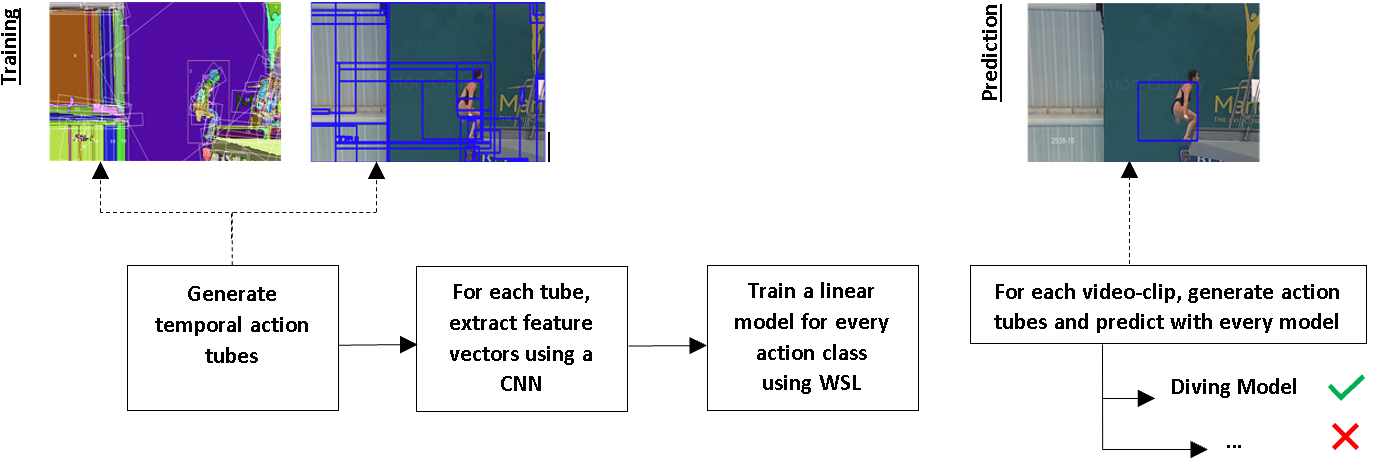}%
\caption{Methodology, high-level overview. Frames from UCF-Sports  \cite{4587727}.}
\label{fig_first_case}
\hfil
\end{figure*}

\section{Methodology}

The proposed action localization methodology is designed to be hierarchical. Specifically, we first take advantage of one of three different techniques to generate temporally consistent video segment proposals, in the form of action tubes. Next, a CNN is used to extract deep features from each action tube. For sake of clarity and generality, once an action tube is vectorised by the CNN, it is  referred to as a {`proposal'}.  Similarly, the originating video is called a  {`set'} (sometimes the term {`bag'} is used in the WSL literature). Generally, each single {`set'} is composed by several proposals. 

Given {`sets'} of proposals derived from the training videos, MIL takes care of selecting (i.e., localizing) for each test set a proposal that best characterizes the action the model represents. This is done by training
a probabilistic model for every action class under a probabilistic \MIL{} formulation, thus taking full advantage of the weakly supervised learning setting.
\\
At prediction time the learned models are employed to localize proposals within the test sets (video-clips), after which one-vs-all classification, using all the learnt class-specific models, is used to infer the set-level label. This label essentially associates a set (i.e., a video-clip) with a specific action class. 

Figure \ref{fig_first_case} provides an overview of this approach. 

\subsection{Generating Action Tube Proposals}

The generation of action tubes from videos can be formally defined by a mapping $F:V \rightarrow \Lambda^3$, where $V$ represents a video and $\Lambda^3$ is a 3D lattice associated with that video. Conversely, a \emph{supervoxel} is a 
subset $\mathcal{V}$ of $\Lambda^3$ representing a group of connected and/or perceptually similar pixels.  

Given an input video $V$, video segmentation produces a set of supervoxels $\{\mathcal{V}_i \subset \Lambda^3,i\}$ such that:
\[
\begin{split}
\bigcup_i \mathcal{V}_i = \Lambda^3 \ \ \text{and} \ \  \mathcal{V}_i \cap \mathcal{V}_j = \emptyset \ \ \forall i,j.
\end{split}
\]
In particular, the video segmentation algorithm proposed by \cite{grundmann2010efficient} first builds a 3D graph from the entire video volume by using image segmentation methods to get an initial over-segmented version of each frame. Subsequently, dense optical flow is used to slice the structure of the graph along the temporal dimension. The author makes use of a hierarchical scheme  to recursively  re-segment the over-segmented frames. This approach enables the algorithm to achieve spatial and temporal cohesive supervoxels even in videos of longer duration. While effective, this algorithm has drawbacks. Firstly, the algorithm does not provide  a method for selecting an optimal hierarchy level (i.e., when should we stop segmenting?). Secondly, the supervoxels that represent human actions still tend to be over-segmented even at the higher levels of the hierarchy. 
\\
To addresses the first issue  \cite{Xu_2013_ICCV} develops a method that makes use of an \textit{`objectiveness'} measure to select the hierarchy level that yields the best spatial and  temporary consistent supervoxels.
The work of \cite{DBLP:journals/corr/WeinzaepfelMS16}  further improves on this idea by developing an algorithm that joins broken supervoxels using a selective search approach.

In this  work we evaluate two spatio-temporal video segmentation  methods for proposal generation: \textit{`Efficient hierarchical graph-based video segmentation (\HVC)'} \cite{grundmann2010efficient} and  \textit{`Action localization with tubelets from motion (\ATM)'} \cite{jain2014action}. Additionally, we evaluate  a recent non-segmentation-based proposal generation method, namely \textit{`Action localization proposals\ from dense trajectories (\ADT)'} \cite{van2015apt}.

\subsection{Proposal Representation}

The \CNN{} from \cite{chatfield2014return} is used in this work to map each proposal to a feature vector. This particular \CNN{} has 16 layers from which fc7 features are extracted for every frame in every action tube. The resulting feature vectors are averaged across all the frames of the constituting action tube and then normalized. This process results in $4096$ feature components per action tube. One side effect of our approach is that since frames are fed to the network separately, no motion-related information is extracted. Neural networks that can learn from an entire video sequence are becoming very popular \cite{tran2014c3d}: their adoption would provide a natural continuation of this work.

\subsection{Weakly Supervised Learning - Training}

Consider a finite set of actions $a \in \mathcal{A}$, each representing a specific action class (e.g., \kicking). For each action class $a$ we define:
$$ 
\mathcal{D}_a = \Big \{ (X_1, Y_{1,a}), \dots, (X_N, Y_{N,a}) \Big \},
$$
where
$X_i$, $i \in \{1, \dots, N\}$, represents a single {`set'} (i.e., a single video) and $Y_{i,a} \in \lbrace 0,1 \rbrace$ is the associated binary set-level label (either the selected action class is present, or it is not). Every set $X_i$ is composed of $j$ proposals (namely, vectorised action tubes) $x_{i,j} \in X_i$, where $j \in \{1, \dots, J\}$. Initially, $Y_{i,a}$ is set to $1$ if $X_i$ contains a proposal that depicts $a$, to $0$ otherwise. 

Each $x_{ij}$ is associated with a latent variable $y_{i,j,a}$, which represents the (unknown) label of the corresponding proposal (video segment). Under the classical \MIL{} assumptions:
\begin{equation}
  Y_{i,a} = \max_j{y_{i,j,a}}.
  \label{eq:y_ia}
\end{equation}
The objective is thus to find, for each action class $a$, a function providing the value of the latent variable for each proposal, namely:
\[
f_a(x_{i,j})  \rightarrow  y_{i,j,a}.
\]

\cite{andrews2002support} formulates the \MIL{} objective function as an instance max-margin problem. This leads to a mixed integer programming problem that can only be solved heuristically.

Additionally, using this formulation prior knowledge cannot be easily incorporated.

For these reasons, here we follow \cite{DBLP:journals/corr/WangZYB15} in-order to relax the MIL constraints such that the latent variable of each proposal is allowed to assume continuous values rather than discrete ones, namely: $y_{i,j,a} \in \lbrack 0,1 \rbrack $. 
\\
The probability, $p_{i,j,a} $ that a specific proposal (action tube) in a set (video-clip) belongs to a given action class is modelled by a logistic function as follows:
\begin{align}
p_{i,j,a} = Pr(y_{i,j,a} = 1 | x_{i,j}; \textbf{w}_a) = \frac{1}{1+\exp{(-\textbf{w}_a^Tx_{i,j})}}.
\label{eq:19}
\end{align}
The probability that set $X_i$ contains at least one proposal depicting $a$ is therefore:
 \begin{align}
  P_{i,a} = Pr(Y_{i,a} = 1 | X_i; \textbf{w}_a) = 1 - \prod_{j=1}^{J}(1 - p_{i,j,a}).
 \label{eq:20}
\end{align}
This completely encapsulates the \MIL{} constraints where a set, $X_i$, is positive only iff at least one proposal, $x_{i,j}$, is positive. Conversely, a set is negative if all instances are negative. The overall optimisation problem is: 

\begin{equation} 
%\begin{split}
\arg\min_{\mathbf{w}_a} \left[ \frac{\lambda}{2} || \mathbf{w}_a ||^2 + \frac{1}{JN}\sum_{i=1}^{N} \left ( J\beta\mathcal{\alpha}_{X_i}  + \sum_{j=1}^{J} \mathcal{\alpha} _{x_{i,j}} \right ) \right],
\label{eq:31}	
\end{equation} 
where
\begin{equation*}
\mathcal{\alpha}_{X_i}  = - \Big [ Y_{i,a} \log P_{i,a} + (1-Y_i) \log(1-P_{i,a}) \Big ]
\end{equation*}
is the cost function for $X_i$, and
\begin{equation*}
\mathcal{\alpha}_{x_{i,j}} = \max \Big ( 0, \eta - \text{sign}(p_{i,j} - \zeta)(\mathbf{w}_a^T {x}_{i,j}) \Big )
\end{equation*}
is the cost function for $x_{i,j}$. 
\\
The following parameters are involved: 
\begin{itemize} 
    \item $\lambda$ is the learning rate.
    \item $\beta$ is a regularization parameter for weight decay. 
    \item $\eta$ is the margin parameter, that separates positives proposals from negative proposals.
    \item $\zeta$ is a threshold parameter that determines whether a proposal should be considered as positive or negative. 
\end{itemize} 
\newcommand{\me}{\mathrm{e}}

Stochastic gradient descent is to used to solve equation \eqref{eq:31}, this dictates the introduction of an additional hyper-parameter $\pi$ that controls the number of iterations per set.

Algorithm \ref{algo::RMISVMTrain} and Algorithm \ref{algo::gd} showcase a pseudo-code implementation of the training function. Note, $\omega$ is exclusively used for set-splitting (discussed below). $\%$ denotes the modulus operator.

\begin{algorithm}[!ht] 
	\SetKwInOut{Input}{Input}
	\SetKwInOut{Output}{Output}
	\textbf{function} TRAIN \\
	\Input{$\{D_a, \mathbf{w}_a, \lambda, \gamma, \beta,\zeta, \eta, \omega, \pi \}$}
	\Output{$\{ \mathbf{w}_a,b_a\}$}
    $\mathcal{T} \leftarrow |D_a| \times \pi$ \\
	\For{$t = 1, ..., T$}
	{
		\If{not $(t \ \ \%  \ \ |D_a|)$} 
		{
		    $D_a$ $\leftarrow$ SPLIT\_SET($D_a$, $\omega$) \\
		    $\pi = \pi-1$ \\
		    return $TRAIN(D_a, \mathbf{w}_a, \lambda, \gamma, \beta, \zeta, \eta, \omega, \pi)$ \\
		}
		$i = t \ \ \%  \ \ |D_a| $ \\
		$X_{i}, Y_{i} \leftarrow D_a[i]$ \ \\
		$p_{i,a} = (1 + \exp(-\mathbf{w}_a^TX_{i}))^{-1}$ \\
		$P_{i,a} = 1- \prod_{j=1}^{J}(1 - p_{i,j,a})$ \\
		$\epsilon = \frac{1}{t \times \lambda}$ \\
        $\alpha = \lbrace r \ \ | sign(p_{i,a}-\zeta) \times \mathbf{w}_a^TX_{i} < \zeta \rbrace $ \\
	    $w_a \leftarrow$ GRADIENT\_DESCENT( $\mathbf{w}_a$,  $X_{i}$, $Y_{i}$, $P_{i,a}$, $p_{i,a}$, $\lambda$, $\beta$, $\zeta$, $\eta$, $\epsilon$) \\
	}
	// split weights and biases \\
	return $\mathbf{w}_a, b_a \leftarrow \mathbf{w}_a$
	\caption{Training}
	\label{algo::RMISVMTrain}	
\end{algorithm}

\begin{algorithm}[!ht]
\SetKwInOut{Input}{Input}
\SetKwInOut{Output}{Output}
\textbf{function} GRADIENT\_DESCENT \\
\Input{$\{\mathbf{w}_a, X_{i}, Y_{i}, P_{i,a}, p_{i,a}, \lambda, \beta, \zeta, \eta\}$}
\Output{$\{\mathbf{w}_a\}$}

$\mathbf{w}_a = (1-\lambda \times \epsilon) \times \mathbf{w}_a$ \\
$\mathbf{w}_a = \mathbf{w}_a + (\beta \times \epsilon \times X_{i} \times \frac{Y_{i}-P_{i,a}}{P_{i,a}} \times p_{i,a}^T)$ \\
$\mathbf{w}_a = \mathbf{w}_a + (\frac{\gamma}{n \times \epsilon \times X_{i}}) . (sign(p_{i,a}-\eta) \times \alpha)^T$ \\
$\mathbf{w}_a = \min \left (  1, \frac{\sqrt{\lambda}^{-1}}{ \sqrt{\sum_{j} w_{a,j}^2}} \right )$\\
return $\mathbf{w}_a$ \\
\caption{Stochastic Gradient Descent}
\label{algo::gd}
\end{algorithm}

\subsubsection{Set Splitting}

\MIL{} captures factors that explain the statistical variations between negative and positive sets. In essence, finding factors of variations that are sufficiently significant to explain the observed difference. Naturally, this makes it difficult to bias MIL formulations towards learning functions that represent only the class of interest and nothing more. Consequently, MIL tends to perform poorly on so called \textit{`hard positives'} or \textit{`hard negatives'}. These are noisy proposals, where determining if the proposal in question is representing the action of interest or not is hard. Geometrically, such proposals lie preciously close to the decision boundaries that separate positives from negatives. For example, differentiating between `running' and `walking' actions.

To mitigate the aforementioned issue, we adopt a \emph{set splitting} technique that leverages the probabilistic formulation of MIL to repeatedly split each set into negatives and positives during training. Specifically, on every epoch,  the proposals $x_{i,j}$ in $X_i$ are sorted according to $p_{i,j,a}$, in descending order. Then the top-$\omega$ proposals $x_{i,j}$ are considered to be positive instances of class $a$, while the rest are deemed to be negative. A new set $X_{i+1}$ is then created to host them (see Algorithm \ref{algo::setsplitting}).

\begin{algorithm}[!ht]
\textbf{function} SPLIT\_SET \\
\SetKwInOut{Input}{Input} 
\SetKwInOut{Output}{Output}
\Input{ $\{D,\omega\}$}
\Output{ $\{D_{new}\}$ }
$D_{new} \leftarrow \emptyset$ \;
\For{$i = 1, ..., |D|$}
{
$X_i, Y_i \leftarrow D[i]$ \\
$X_{positive,i} = \lbrace x | x \in X_i \land  y \in Y_i \land y > \omega \rbrace$  \\
$Y_{positive,i} =  \lbrace y | y \in Y_i \land y > \omega \rbrace$ \\
$X_{negative,i} = X_i \setminus 
X_{positive,i}$ \\
$Y_{negative,i}= \lbrace 0 | x \in X_i \rbrace$ \\
$D_{new} = D_{new} \cup (X_{negative,i},Y_{negative,i}) \cup (X_{positive,i},Y_{positive,i}) $ \\
}\
return $D_{new}$
\caption{Set Splitting}
\label{algo::setsplitting}
\end{algorithm}

\subsection{Weakly Supervised Learning - Predictor}

Given a model 
$f(X_i;\mathbf{w}_a; {b}_a)$
representing a single action class $a$, proposal probabilities are given by equation \eqref{eq:19}. Each probability, $p_{i,j,a}$ is interpreted as a measure of how likely it is for a given proposal, $x_{i,j,a}$ to depict action class $a$. This answers the localisation problem, where the proposal with the highest probability is considered to be the most likely proposal to depict $a$ in $X_i$. 

The probability that set $X_i$ contains a proposal that depicts $a$ is derived from  proposal probabilities by equation \eqref{eq:20}. Set probabilities are used to resolve the classification problem where we would like to associate each set (video-clip) with a single action-class. For this purpose, one-vs-all classification is used.

Algorithm \ref{algo::RMISVMPredict} outlines the prediction function. This predicts both set and proposal probabilities for every new observation, with every trained action class model.  

\begin{algorithm}[!ht]
    \textbf{function} PREDICTOR \\
	\SetKwInOut{Input}{Input}
	\SetKwInOut{Output}{Output}
	\Input{$D_a,\; \mathbf{w}_a,\; b_a$}
	\Output{$\textrm{set\_probabilities}$, $\textrm{proposal\_probabilities}$}
	$\textrm{set\_probabilities} = [ ]$  // list  \\
	$\textrm{proposal\_probabilities} = [ ]$ // list \\
	\For{$i = 1, ..., |D_a|$}
	{
		$X_{i} \leftarrow D_a[i]$ \\
		$p_{i,a} = (1 + \exp(-\mathbf{w}_a^TX_{i}))^{-1}$ \\
		$P_{i,a} = 1 - \prod_{j=1}^{J}(1 - p_{i,j,a})$ \\
		$\textrm{proposal\_probabilities.append}(p_{i,a})$ \\
		$\textrm{set\_probabilities.append}(P_{i,a})$ \\
	}
	return $\textrm{set\_probabilities}$, $\textrm{proposal\_probabilities}$
	\caption{Predictor}
	\label{algo::RMISVMPredict}	
\end{algorithm}

\begin{figure}[h!]
	\centering
	\includegraphics[width=0.75\columnwidth]{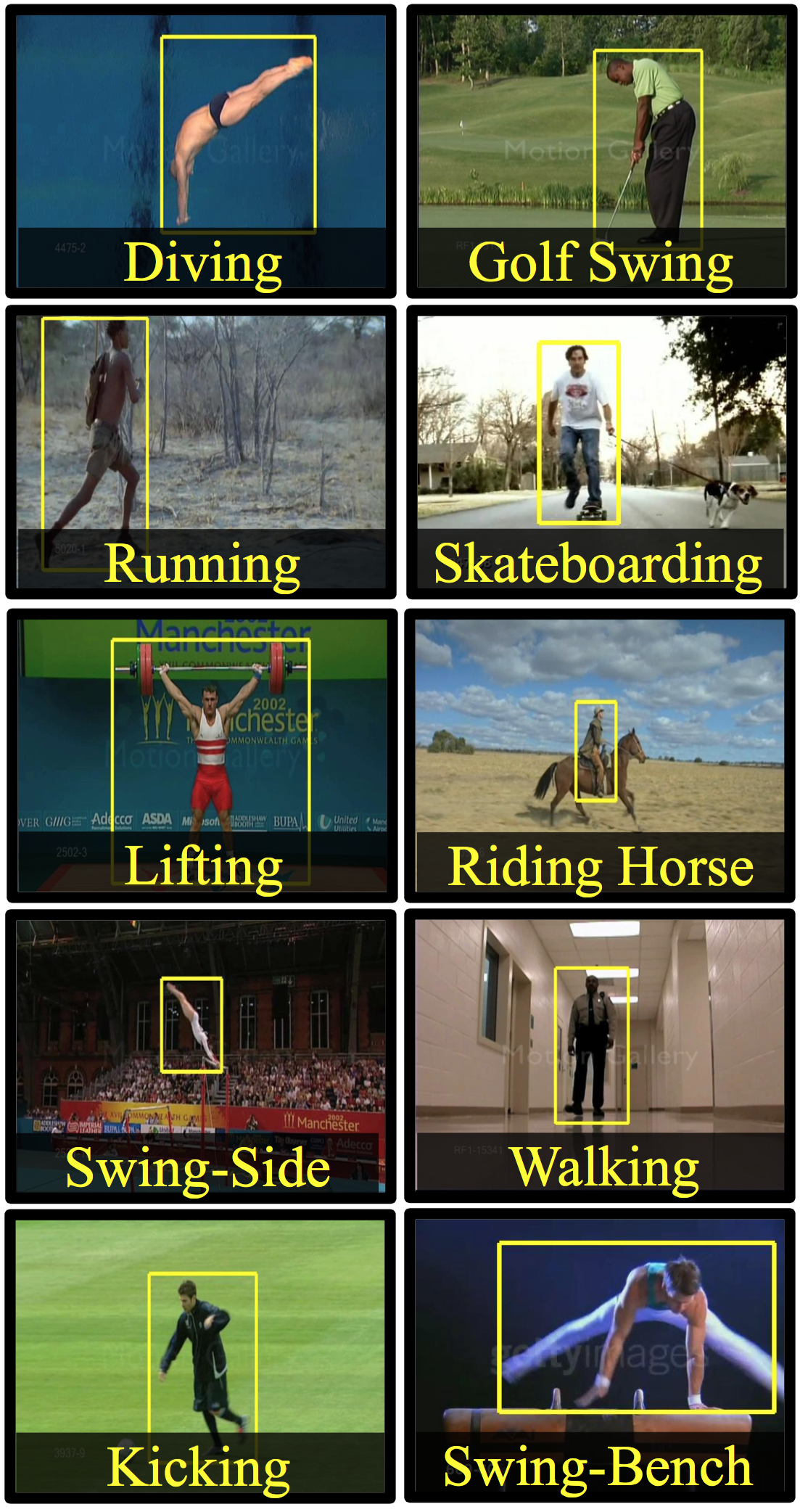}
	\caption{Action-classes from UCF-Sports.}
	\label{fig:WSLSegCurve1}
\end{figure}

\begin{figure*}[t]
	\centering
	\includegraphics[width=2\columnwidth]{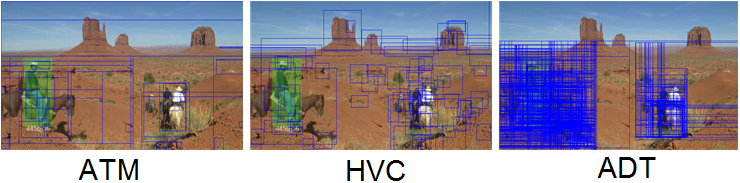}%
	\caption{Proposals (represented by blue bounding-boxes) generated by \ATM, \HVC, \ADT, respectively, on the same example frame. The green overlay depicts the ground-truth.}
	\label{IOURenderings}
\end{figure*}

\section{Experimental Setup}

\subsection{Benchmark and Protocols}

The UCF-Sports data-set is used to evaluate our approach. It consists of one hundred and fifty videos divided into ten separate action classes. Whereas this data-set comes with both set (video) and proposal (frame) level annotations, our approach does not require proposal-level annotations and as such they are only used for evaluation purposes. With regards to the train-test split we follow the approach of \cite{lan2011discriminative} where forty three videos are used for testing and hundred and three videos for training. For cross-validation a slightly modified version of the leave-one-out method is used. Namely, instead of leaving only one observation out, we leave one out for every class while still guaranteeing that every class is in the training set at least once. We make use of a standard grid-search for selecting  hyper-parameter values, optimizing only for set-level classification accuracy.

As for proposal generation, we evaluate and compare three algorithms (\ATM, \HVC, \ADT), from which we select one for the subsequent stages. 

\subsection{Experiments}

After extracting features using the selected proposal generation algorithm we train a probabilistic model for every action class, as described above. 
We set up three different experiments: 

\begin{itemize} 
    \item \textbf{PMIL}: baseline probabilistic MIL;
    \item \textbf{PMIL+F}: probabilistic MIL with filtering, where exceedingly large proposals are filtered-out;
    \item \textbf{PMIL+F+S}: probabilistic MIL with filtering and splitting. 
\end{itemize} 
For each experiment we train a new set of models and evaluate and compare both set-level classification and localization performance. In general, an optimal model should pick a proposal that perfectly encapsulates the action it represents while simultaneously providing accurate set-level classification performance.

\subsubsection{Performance Indicators}

We employ the Intersection Over Union (\IOU) metric to evaluate the quality of the generated proposals with respect to the ground-truth. To evaluate set-level classification and localization performance we follow the lead of previous work \cite{sapienzalearning, tang2013discriminative} by using Mean Average Precision (\mAP). For the purposes of calculating these scores, as in previous work we consider any proposal that covers at least 20\% of the ground truth to be correct. 

In contrast to other work, we measure the localization performance without considering classification accuracy. This metric is  referred to as \textit{Mean-Squared Error with Respect to Optimal Choice} or in short, \mSERO.
\\
This score is given, for each action class $a$ by:
\begin{align}
 \mSERO_{a} = \frac{1}{N} \sum_{i=1}^{N} (k_{i,a} - \max_j{p_{i,j,a}})^2,
\label{eq:56}
\end{align}
where $k_{i,a}$ is the probability assigned to the optimal choice (best possible proposal as measured by \IOU). According to this metric an optimal model will always pick the proposal with the best \IOU, whereas a sub-optimal model will not. This measure quantifies how far off a  model was from predicting the best possible proposal as generated by the first stage of our pipeline. 
\begin{table}[h!]
	\centering
	\caption{Intersection Over Union (\IOU) for UCF-Sports.}
	\label{tab:VSScores} 
	\resizebox{\columnwidth}{!}{
	\begin{tabular}{ l | l l l}
		\hline
		\textbf{Action Class} & \textbf{\ADT} \cite{van2015apt} & \textbf{\ATM} \cite{jain2014action} & \textbf{\HVC} \cite{grundmann2010efficient} \tabularnewline
		\hline
		Diving-Side & \textbf{0.583706}& 0.363168 & 0.452065\tabularnewline
		Golf-Swing-Back & \textbf{0.801282} & 0.748542 & 0.436035\tabularnewline
		Golf-Swing-Front & 0.435979 & 0.526266 & 0.533355\tabularnewline
		Golf-Swing-Side & 0.724060 & \textbf{0.727153} & 0.461906\tabularnewline
		Kicking-Front & \textbf{0.696297} & 0.550503 & 0.337375\tabularnewline
		Kicking-Side & \textbf{0.691274} & 0.484475 & 0.498129\tabularnewline
		Riding-Horse &\textbf{0.542712} & 0.247956 & 0.467046\tabularnewline
		Run-Side & \textbf{0.588205} & 0.492336 & 0.351405\tabularnewline
		SkateBoarding-Front &\textbf{0.659791 }& 0.561727 & 0.319265\tabularnewline
		Swing-Bench & \textbf{0.683738}& 0.464019 & 0.400841\tabularnewline
		Swing-SideAngle & \textbf{0.529371} & 0.368790 & 0.359946\tabularnewline
		Walk-Front & \textbf{0.713729} & 0.640201 & 0.425863\tabularnewline
		\hline
		\hline
		\textbf{Mean \textsc{IOU}} & \textbf{0.637512} & 0.514595 & 0.420269\tabularnewline
		\hline
		\textbf{Mean \#Proposals} & 1798 & \textbf{254} & 312 \tabularnewline
		\hline
		\hline
	\end{tabular}	}
\end{table}

\section{Results}

\subsection{Proposal Generation}

\begin{figure} [h!]
	\centering
	\includegraphics[width=\columnwidth]{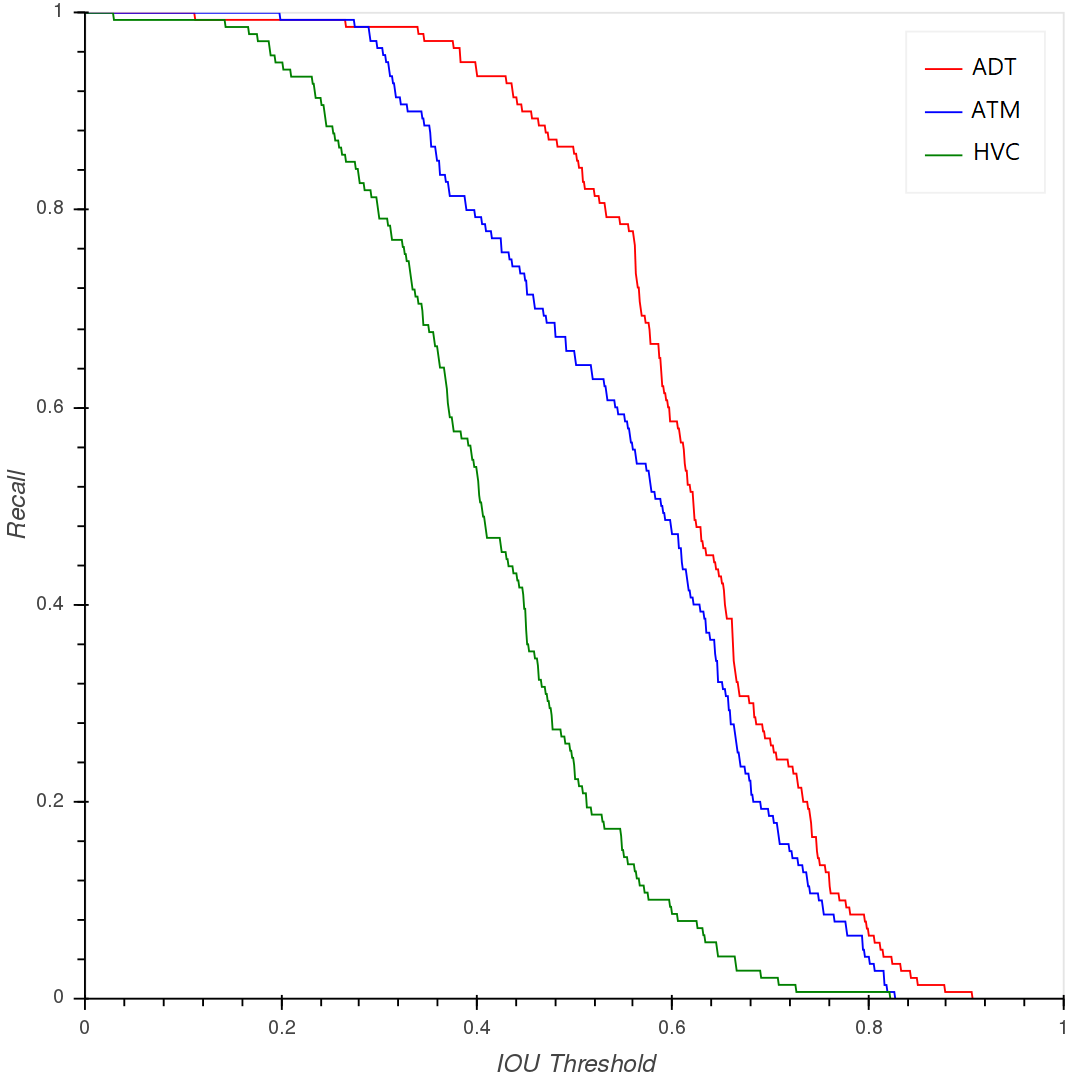}
	\caption{Recall-\IOU{} Threshold curve computed on UCF-Sports with three different proposal generation algorithms.}
	\label{fig:RECALLIOU}
\end{figure}

Table \ref{tab:VSScores} reports the \IOU{} score of the best action tube for every action class, averaged across all the videos in that class. Figure \ref{fig:RECALLIOU} depicts the Recall-\IOU{} curve at various thresholds. Figure \ref{IOURenderings} illustrates a single frame from the \textit{`horse-riding'} action-class in UCF-Sports rendered with the generated action tube proposals  alongside the ground-truth.

The results clearly show how the \ADT{} algorithm achieves an overall better \IOU{} then the other algorithms. However, as it can be observed in Figure \ref{IOURenderings}, it also generates significantly more proposals. In fact \ADT{} generates approximately seven proposals for each proposal generated by \ATM{} or \HVC. An excessive number of proposals per video is not desirable, since it makes the localization task harder, negatively effecting generalization performance. Primarily for this reason we decided to use the proposals generated by the \ATM{} algorithm for feature extraction. \ATM{} provides a reasonable balance between the number of proposals generated and the quality of the resulting video segments.

\subsection{Classification and Localization}

Table \ref{tab:MAPScores} lists the \mAP{} scores for all action classes generated in the aforementioned experiments.

\begin{table}
	\small
	\caption{\mAP{} scores for UCF-Sports (\mAP{}/\AP @ 0.2).}
	\label{tab:MAPScores} 
	\resizebox{\columnwidth}{!}{
	\begin{tabular}{l | l | l | l }
		\hline
		\textbf{CLASS} & \textsc{\textbf{PMIL}} &\textsc{\textbf{PMIL+F}} & \textsc{\textbf{PMIL+F+S}} \tabularnewline
		\hline
		kicking & 64.00  & 64.00   & 64.00  \tabularnewline
		golf-swing & 79.46  & 79.46  &  79.46 \tabularnewline
		diving & 100.00  & 100.00    & 100.00  \tabularnewline
		riding-horse & 12.50 & 12.50   & 50.00  \tabularnewline
		running & 0  & 42.00   & 50.01 \tabularnewline
		skate-boarding &  0  &  43.01  & 50.08  \tabularnewline
		swing-bench &   100.00  & 100.00    & 100.00  \tabularnewline
		swing-side & 0.00  & 56.25   & 56.25  \tabularnewline
		walking & 26.84 & 40.16  & 40.16 \tabularnewline
		\hline	
		\textbf{\mAP{}} & 43.00  &  60.00  &  67.00 \tabularnewline	
		\hline	
		\hline		
	\end{tabular}}
\end{table}

One can observe that filtering the data to exclude proposals that have high intersection with the entire video volume delivers significant improvements (PMIL+F). The reason can be attributed to the fact that large proposals tend to be `hard negatives', implying that the learner does not have an incentive to pick a tight-fitting proposal over a larger proposal, assuming both proposals adequately cover the action in question. It can also be seen that the video splitting technique also has a positive effect (PMIL+F+S). Both observations empirically validate our conjecture that MIL-based techniques generalize  well, as long as proposal ambiguity is minimized. 

\begin{figure}  
	\centering
	\includegraphics[width=1\columnwidth]{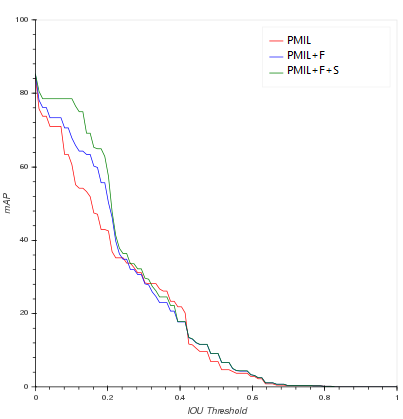}
	\caption{Plot of \mAP{} score versus \IOU{} for all experiments.}
	\label{fig_first_case1}
\end{figure}

Figure \ref{fig_first_case1} plots the \mAP{} score versus the \IOU{} threshold. As expected, increasing the threshold parameter causes the \mAP{} score to decrease significantly. Recall that when the threshold is increased the scoring algorithm is forced to be more selective as to what proposals it considers correct, thereby increasing the classification error. 

\subsubsection{\mSERO}

Table \ref{tab:MSERO} lists the \mSERO{} scores for all action class in the considered experiments. Lower values imply that a proposal that is close to the optimal one has been selected. On the contrary, higher values indicate that the model selected a bad proposal.Recall, an optimal proposal is one that has high \IOU{} relative to the ground truth. The best performing class-specific models are those for \textit{`diving'} followed closely by \kicking and \swinging. The effect of set-splitting (PMIL+F+S) on localization performance is significant.

\begin{table} 
	\small
	\caption{\mSERO{} scores for UCF-Sports.}
	\label{tab:MSERO} 
	\resizebox{\columnwidth}{!}{
	\begin{tabular}{l | l | l | l }
		\hline
		\textbf{CLASS} & \textbf{PMIL} & \textbf{PMIL+F}  & \textbf{PMIL+F+S} \tabularnewline
		\hline
		kicking & 4.410  & 3.771 & 3.771 \tabularnewline
		golf-swing & 4.868  & 4.606  &  4.606 \tabularnewline
		diving & 0.429 & 1.216   & 1.216  \tabularnewline
		riding-horse & 3.323 & 3.456 & 1.523  \tabularnewline
		running & 10.152  & 10.180  & 0.659 \tabularnewline
		skate-boarding &  12.228  &  12.228  &  12.228  \tabularnewline
		swing-bench &   5.450 & 4.644   & 4.644  \tabularnewline
		swing-side & 11.082  & 9.456   & 9.456  \tabularnewline
		walking & 8.785 & 6.969  & 6.969 \tabularnewline
		\hline	
		\hline		
	\end{tabular}}
\end{table}

\begin{figure}[h!]
	\centering
	\subfloat[Skateboarding Model - UCF-Sports Skateboarding video-clip\#001 ]{\includegraphics[width=1\columnwidth]{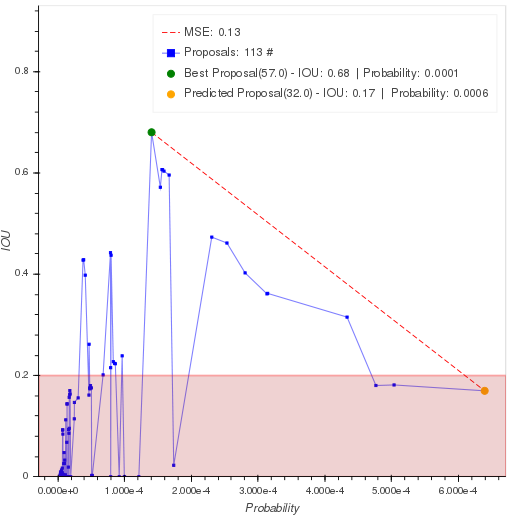}
	\label{fig_first_case12}}
	\hfil
	\subfloat[Diving Model - UCF-Sports Diving video-clip\#004]{\includegraphics[width=1\columnwidth]{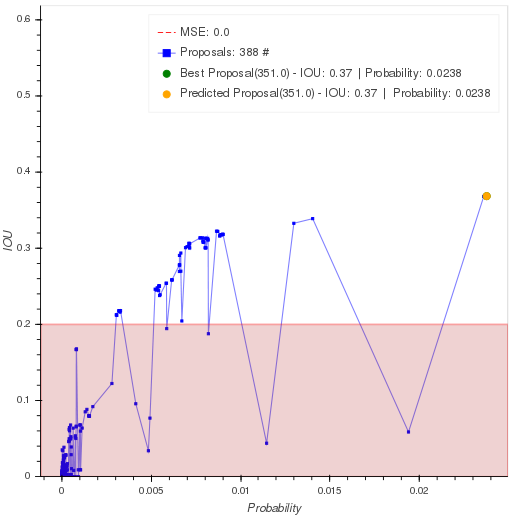}
	\label{fig_second_case22}}
	\caption{Proposal scatter plots, depicting the relationship between \IOU{} and the localization probability assigned to every proposal from two  videos clips, \textit{`skateboarding'} (a) and \textit{`diving'} (b).}
	\label{fig_sim2}
	\vspace{-0.5cm}
\end{figure}

Figure \ref{fig_sim2} examines the localization performance of two different models. Each figure plots all the proposals from a single video, highlighting the exact probability assigned by our model versus the actual \IOU{} with respect to the ground truth. The proposal that actually selected (orange circle) by the model as well as the optimal proposal (green circle) are highlighted. The red section depicts the minimum amount of \IOU{} (20\%) a selected proposal has to exceed in-order for it be considered `correct'. From these graphs one can easily see how the best performing model (i.e., that for \textit{`diving'}) assigns a high probability to the proposal closer to the optimum (in terms of \IOU{}). On the other hand, a sub-optimal model (such as that for \textit{`skateboarding'}) assigns a high probability to a proposal with a low \IOU{}.  

\subsubsection{Qualitative Discussion}

\begin{figure}[t]
  \centering
  \begin{tabular}{@{}c@{}}
    \includegraphics[width=3in]{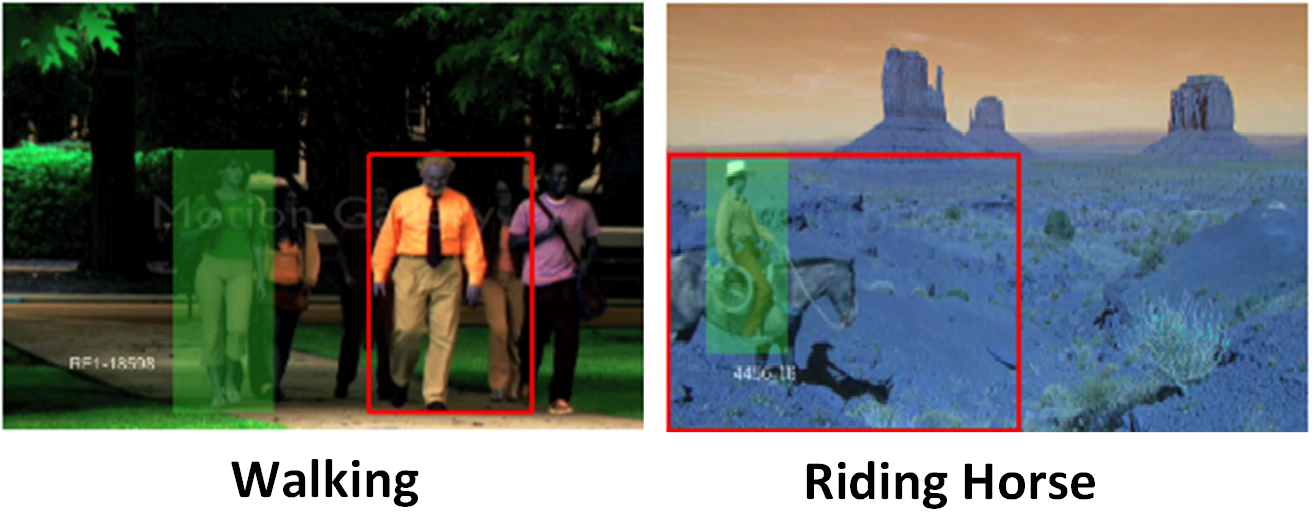} \\[\abovecaptionskip]
    \small (a) Worst-Case  localisation.
    \label{lola}
  \end{tabular}
  \vspace{\floatsep}
  \begin{tabular}{@{}c@{}}
    \includegraphics[width=3in]{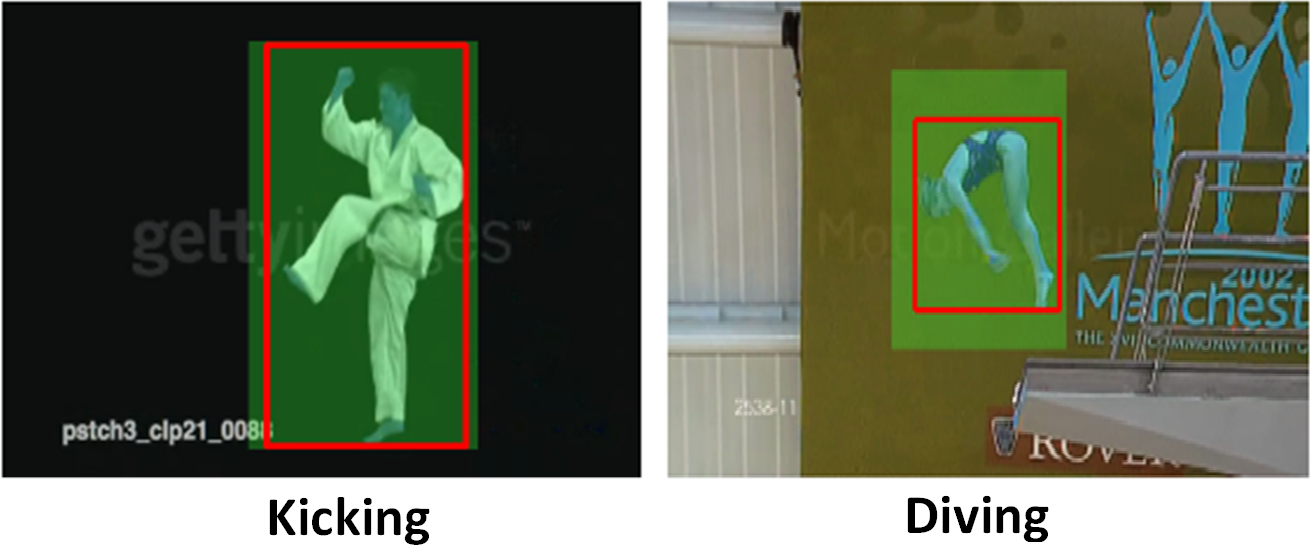} \\[\abovecaptionskip]
    \small (b) Best-Case localisation.
    \label{lol2a}
  \end{tabular}
  \caption{Examples of worst-case (a) and best-case (b) localization results. The selected proposals are visualised as red bounding boxes. The ground-truth is denoted by a green bounding box.}
  \label{fig:myfig}
\end{figure}

Table \ref{table:results} compares the results achieved by our methodology with that of \cite{tang2013discriminative}. The results clearly demonstrate the effectiveness of our approach (note that \cite{tang2013discriminative} does not breakdown scores on per class basis). Figure \ref{fig:myfig} illustrates successfully and unsuccessfully localised action instance of the UCF-Sports data-set. Interestingly, the frame showcasing the \textit{`walking'} action  depicts three human bodies who appear to be walking. However, the provided ground truth annotation only covers one of these instances of the walking action. Our model picked the proposal covering the most salient body. Objectively, this is correct answer. Similarly, in the  \textit{`diving'} frame one can argue that the prediction given by our model is even better then the one provided by the ground-truth. Other similar examples can be provided.

\begin{table}  
	\centering
	\caption{UCF-Sports result comparison.}
	\begin{tabular}{l|l|l}
		\hline
		\textbf{Class} & \textbf{Ours} &  \cite{tang2013discriminative}  \\
		\hline
		kicking & 64.00 & \_  \\
		golf-swing & 79.46 & \_ \\
		diving & 100.00 & \_  \\
		riding-horse &  50.00 & \_  \\
		running & 50.08 & \_ \\
		skate-boarding & 50.00 & \_  \\
		swing-bench & 100 & \_  \\
		swing-side & 56.25 & \_ \\
		walking & 40.16 & \_ \\
		\hline
		\textbf{\mAP{}} &  \textbf{67.00} & 61.20  \\    	  \hline
		\hline
	\end{tabular}
	\label{table:results}
\end{table}

\section{Conclusion}

In this work we proposed a novel framework for the localization and classification of space-time actions under weakly labelled conditions. Experimental results on the UCF-Sports data-set prove the effectiveness of our approach. In summary,  we first generate temporally consistent proposals. Then a \CNN is used to  extract \RGB{} features from each proposal. Finally, we use a set-splitting technique to reduce ambiguity between different proposals and employ a probabilistic formulation of \MIL. 
The latter serves to make the resulting optimization function more tractable while also enabling the integration of prior knowledge. While we leave this for future work, this  can be seen as stepping stone in that direction.  

\section*{Acknowledgements}
This work was in part supported by the Malta ENDEAVOUR scholarship scheme. The work was also partly supported by the European Union’s Horizon 2020 research and innovation programme under grant agreement No. 779813 (SARAS).

\bibliographystyle{plain}
\bibliography{ref} 

\end{document}